\documentclass[11pt,a4paper]{article}
\usepackage[hyperref]{acl2020}
\usepackage{times}
\usepackage{latexsym}

\usepackage{multirow}
\usepackage{amsmath}
\usepackage{mathtools}
\usepackage{graphicx}
\usepackage{algorithm}
\usepackage{balance}
\usepackage{microtype}

\aclfinalcopy

\usepackage{amsfonts, amssymb}
\usepackage{multirow}
\usepackage{algpseudocode}

\algnewcommand{\IIf}[1]{\State\algorithmicif\ #1\ \algorithmicthen}
\algnewcommand{\EndIIf}{\unskip\ \algorithmicend\ \algorithmicif}

\title{Relabel the Noise: Joint Extraction of Entities and Relations via Cooperative Multiagents}

\author{
Daoyuan Chen\textsuperscript{1}~~~
Yaliang Li\textsuperscript{1}~~
Kai Lei\textsuperscript{2}~~
Ying Shen\textsuperscript{3} \thanks{~~~Corresponding author.} \\
\textsuperscript{1}{Alibaba Group} ~~ \\
\textsuperscript{2}{School of Electronics and Computer Engineering, Peking University} \\\textsuperscript{3}{School of Intelligent Systems Engineering, Sun Yat-Sen University}\\
\texttt{\textsuperscript{1}\small  \{daoyuanchen.cdy,~yaliang.li\}@alibaba-inc.com} \\\texttt{\textsuperscript{2}\small leik@pkusz.edu.cn,} 
\texttt{\textsuperscript{3}\small sheny76@mail.sysu.edu.cn}
}

\begin{document}
\maketitle

\begin{abstract}
Distant supervision based methods for entity and relation extraction have received increasing popularity due to the fact that these methods require light human annotation efforts. 
In this paper, we consider the problem of \textit{shifted label distribution}, which is caused by the inconsistency between the noisy-labeled training set subject to external knowledge graph and the human-annotated test set, and exacerbated by the pipelined entity-then-relation extraction manner with noise propagation.
We propose a joint extraction approach to address this problem by re-labeling noisy instances with a group of cooperative multiagents.
To handle noisy instances in a fine-grained manner, each agent in the cooperative group evaluates the instance by calculating a continuous confidence score from its own perspective;
To leverage the correlations between these two extraction tasks, a confidence consensus module is designed to gather the wisdom of all agents and re-distribute the noisy training set with confidence-scored labels. 
Further, the confidences are used to adjust the training losses of extractors.
Experimental results on two real-world datasets verify the benefits of re-labeling noisy instance, and show that the proposed model significantly outperforms the state-of-the-art entity and relation extraction methods. 
\end{abstract}

\section{Introduction}

\begin{figure*}[t]
\centering
\includegraphics[width=0.91\linewidth]{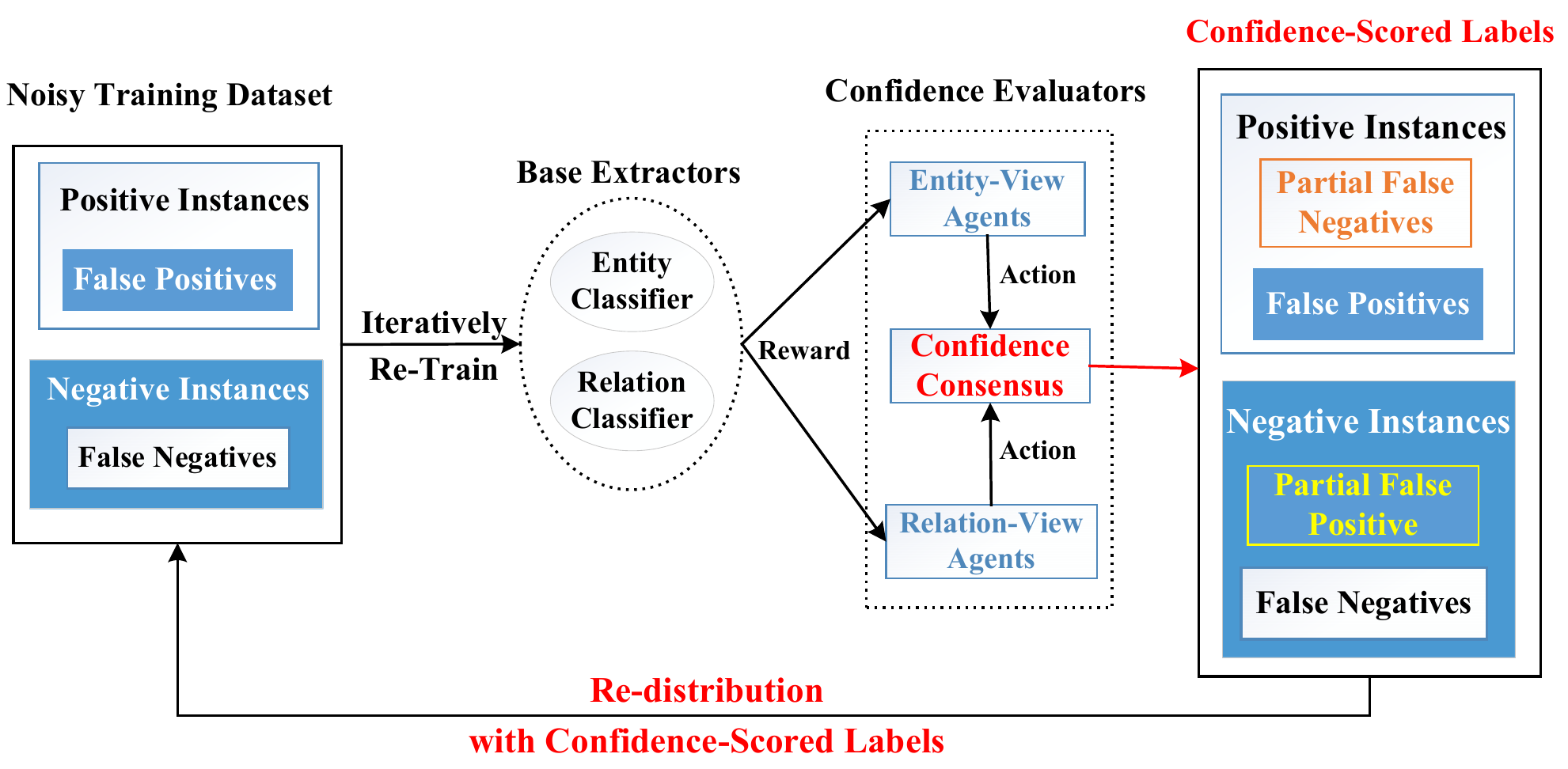}
\caption{Overview of the proposed method. A group of multiagents are leveraged to evaluate the confidences of noisy instances from different extraction views. 
Base extractors are refined by iteratively training on the re-distributed instances with confidence-scored labels.
}
 \label{fig-overview}
\end{figure*}

The extraction of entities and relations has long been recognized as an important task within natural language processing, as it facilitates text understanding.
The goal of the extraction task is to identify entity mentions, assign predefined entity types, and extract their semantic relations from text corpora. 
For example, given a sentence \textit{``Washington is the president of the United States of America"}, an extraction system will find a \textsc{president\_of} relation between \textsc{person} entity ``\textit{Washington}'' and \textsc{country} entity ``\textit{United States of America}''.

A major challenge of the entity and relation extraction task is the absence of large-scale and domain-specific labeled training data due to the expensive labeling efforts.
One promising solution to address this challenge is distant supervision (DS) \cite{mintz2009distant,hoffmann2011knowledge}, which generates labeled training data automatically by aligning external knowledge graph (KG) to text corpus.
Despite its effectiveness, the aligning process introduces many noisy labels that degrade the performance of extractors. 
To alleviate the introduced noise issue of DS, extensive studies have been performed, such as using probabilistic graphical models \cite{surdeanu2012multi}, neural networks with attention \cite{zeng2015distant,lin2016neural} and instance selector with reinforcement learning (RL) \cite{acl2018qinRL,feng2018RLRE}.

However, most existing works overlooked the \textbf{shifted label distribution} problem \cite{ye2019looking}, which severely hinders the performance of DS-based extraction models.
Specifically, there is a label distribution gap between DS-labeled training set and human-annotated test data, since two kinds of noisy labels are introduced and they are subject to the aligned KG: 
(1)	False Positive: unrelated entity pair in the sentence while labeled as relations in KG; and 
(2)	False Negative: related entity pair while neglected and labeled as \textsc{NONE}.
Existing denoising works assign low weights to noisy instances or discard false positives while not recovering the original labels, leaving the shifted label distribution problem unsolved.

Moreover, most denoising works assume that the target entities have been extracted, i.e., the entity and relation extraction is processed in a pipe-lined manner. By extracting entities first and then classifying predefined relations, the entity extraction errors will be propagated to the relation extractor, introducing more noisy labels and exacerbating the shifted label problem.
Besides, there are some correlations and complementary information between the two extraction tasks, which are under-utilized but can provide hints to reduce noises more precisely, e.g., it is unreasonable to predict two \textsc{country} entities as the relation \textsc{president\_of}.

In this paper, to reduce the shifted label distribution gap and further enhance the DS-based extraction models, we propose a novel method to re-label the noisy training data and jointly extract entities and relations. 
Specifically, we incorporate RL to re-label noisy instances and iteratively re-train entity and relation extractors with adjusted labels, such that the labels can be corrected by trial and error.
To leverage the correlations between the two extraction tasks, we train a group of cooperative multiagents to evaluate the instance confidence from different extraction views. 
Through a proposed confidence consensus module, the instances are re-labeled with confidence-scored labels, and such confidence information will be used to adjust the training loss of extractors. 
Finally, the performances of extractors are refined by exploring suitable label distributions with iterative re-training.

Empirical evaluations on two real-world datasets show that the proposed approach can effectively help existing extractors to achieve remarkable extraction performance with noisy labels, and the agent training is efficient with the help of correlations between these two extraction tasks.

\section{Methodology}
\subsection{Overview}
In this research, we aim to refine entity extractor and relation extractor trained with DS, by incorporating a group of cooperative multiagents. 
Formally, given a DS training corpus $D=\{s_1, \dots, s_n\}$, an entity extractor $\theta_e'$ and a relation extractor $\theta_r'$ trained on $D$ are input into the multiagents. 
The agents re-distribute $D$ with confidence-scored labels and output two refined extractors $\theta_e^*$ and $\theta_r^*$ using the adjusted labels.

Towards this purpose, we model our problem as a decentralized 
multiagents RL problem, where each agent receives local environmental observation and takes action individually without inferring the policies of other agents.
It is hard to directly evaluate the correctness of adjusted noisy labels since we do not know the ``gold'' training label distributions suitable to the test set. 
Nonetheless, we can apply RL to indirectly judge the re-labeling effect by using performance scores on an independent validation set as rewards, which is delayed over the extractor re-training.
Further, the decentralization setting allows the interaction between the distinct information of entity and relation extractors via intermediate agents.

As shown in Figure \ref{fig-overview}, a group of agents acts as confidence evaluators, and the external environment consists of training instances and classification results of extractors. 
Each agent receives a private observation from the perspective of entity extractor or relation extractor, and makes an independent action to compute a confidence score of the instance. 
These actions (confidence scores) will then be considered together by the confidence consensus module, which determines whether the current sentence is positive or negative and assigns a confidence score. 
Finally, the updated confidences are used to retrain extractors, the performance score on validation set and the consistent score of the two extractors are combined into rewards for agents.

The proposed method can be regarded as a post-processing plugin for existing entity and relation extraction model. 
That is, we design a general framework of the states, actions and rewards by reusing the inputs and outputs of the extractors.

\subsection{Confidence Evaluators as Agents}
\label{sec:agents}
A group of cooperative multiagents are used to evaluate the confidence of each instance. 
These multiagents are divided into two subgroups, which act from the perspective of entity and relation respectively.
There can be multiple agents in each subgroup for the purpose of scaling to larger observation space and action space for better performance. 
Next, we will detail the states, actions and rewards of these agents.

\paragraph{States}
The states $S_e$ for entity-view agents and $S_r$ for relation-view agents represent their own viewpoint to evaluate the instance confidence. 
Specifically, entity-view agents evaluate sentence confidence according to three kinds of information: current sentence, the entity extraction results (typed entity) and the noisy label types.
Similarly, relation-view agents make their decisions depending on the current sentence, the relation types from relation extractor and the noisy label types from DS.

Most entity and relation extractors encode the semantic and syntactic information of extracted sentences into low-dimension embeddings as their inputs.
For entity types and relation types, we also encode them into embeddings and some extractors have learned these vectors such as CoType \cite{ren2017cotype}.
Given reused extractors, we denote the encoded sentence vector as $\bf{s}$, the extracted type vector as $\bf{t^e}$ and $\bf{t^r}$ for entity and relation respectively, and DS type vectors as $\bf{t_d^e}$ and $\bf{t_d^r}$ for entity and relation respectively. 
We reuse the sentence and type vectors of base extractors to make our approach lightweight and pluggable. 
Finally, we average the extracted and DS type embeddings to decrease the size of observation space, and concatenate them with the sentence embedding $\bf{s}$ to form the states $\bf{S^e}$ and $\bf{S^r}$ for entity/relation agents respectively as follows:
\begin{equation}
\label{equ:states}
\bf{
S^e =  s \Vert (t^e+t_d^e)/2, \quad  S^r = s \Vert (t^r+t_d^r)/2,
}
\end{equation}
Note that we have encoded some semantics into the type vectors, e.g., the margin-based loss used in CoType enforces the type vectors are closer to their candidate type vectors than any other non-candidate types. 
Intuitively, in the representation spaces, the average operation leads in the midpoint of extracted type vector and DS type vector, which partially preserves the distance property among the two vectors and other type vectors, so that helps form distinguishable states. 

\paragraph{Actions}
To assign confidence in a fine-grained manner and accelerate the learning procedure, we adopt a continuous action space.
Each agent uses a neural policy network $\Theta$ to determine whether the current sentence is positive (conform with the extracted type $t_i$) or negative (``None" type) and computes a confidence score $c$.
We model this action as a conditional probability prediction, i.e., estimate the probability as confidence given by the extracted type $t_i$ and the current state $\bf{S}$: $c=p (positive|t_i,\Theta,\bf{S})$.
We adopt gated recurrent unit (GRU) as policy network, which outputs the probability value using sigmoid function.
A probability value (confidence score) which is close to 1/0 means that the agent votes a sentence as positive/negative with a high weight.

To handle huge state spaces (e.g., there are thousands of target types in our experimental dataset) and make our approach scalable, here we divide and conquer the state space by using more than one agent in entity-view and relation-view groups. 
The target type set is divided equally by agent number and each agent only is in charge of a part of types.
Based on the allocation and DS labels, one sentence is evaluated by only one relation agent and two entity agents at a time, meanwhile, the other agents are masked.

\paragraph{Re-labeling with Confidence Consensus}
\label{sec:consensus}
To leverage the wisdom of crowds, we design a consensus strategy for the evaluated confidences from multiagents.
This is conducted by two steps: gather confidences and re-label with confidence score.
Specifically, we calculate an averaged score as $\bar c = c_{sum}/3$, where $c_{sum}$ is the sum of all agent confidences and the dividing means three agents evaluated the present sentence due to the above masking action strategy.
Then we label the current sentence as negative (``None" type) with confidence $C=1-\bar c$ if $\bar c \leq 0.5$, otherwise we label the current sentence as positive (replace noisy label with extracted type) with confidence $C=\bar c$.
This procedure can be regarded as weighted voting and re-distribute the training set with confidence-scored labels as shown in the right part of Figure \ref{fig-overview}, where
some falsely labeled instances are put into intended positions or assigned with low confidences.

\paragraph{Rewards}
The reward of each agent is composed of two parts: shared global reward $g$ expressing correlations among sub-tasks, and separate local rewards restricting the reward signals to different three agents for different sentences (recall that we evaluate each sentence by different agents w.r.t their responsible types).
Specifically, the global reward $g$ can give hints for denoising and 
here we adopt a general, translation-based triple score as used in TransE \cite{bordes2013translating}
$g=||\bf{t_1+t_r-t_2}||$, where $\bf{t_1}$, $\bf{t_r}$ and $\bf{t_2}$ are embeddings for triple $(E_1, R, E_2)$ and pre-trained by TransE.
The score is used to measure the semantic consistency of each triple and can be easily extended with many other KG embedding methods \cite{wang2017knowledge}.
As for the separate local reward, we use F1 scores $F_1^e$ and $F_1^r$ to reflect the extractor performance, which are gained by entity extractor and relation extractor on an independent validation dataset \footnote{To gain a relatively clean data, we randomly select 20\% data from the original training set, extract them using pre-trained CoType model and retain only one instance for each sentence whose DS label is the same as the extracted label.}
respectively.
Finally, to control the proportions of two-part rewards, we introduce a hyper-parameter $\alpha$, which is shareable for ease of scaling to multiple agents as: 
\begin{equation}
\label{equ:reward}
    r_e = \alpha*F_1^e - g, \quad  r_r = \alpha*F_1^r - g.
\end{equation}

\subsection{Model Learning}
\label{sec:training}
\subsubsection{Loss Correction for Extractors}
With the evaluated confidences and re-labeled instances, we adjust the training losses of entity extractor and relation extractor to alleviate the performance harm from noise and shifted label distribution.
Denote the original loss of extractor as $\ell$, the new loss $\ell'$ is adjusted by an exponential scaling factor $\lambda$ and confidence $C$ as : $\ell' = C^\lambda\ell$.
Intuitively, a small confidence score $C$ and a large $\lambda$ indicate that the current instance has almost no impact on the model optimization. 
This can alleviate side-effects caused by noises and prevent the gradient being dominated by noisy labels, especially for those with divergent votes since the averaging in confidence consensus module leads to a small $C$.

\subsubsection{Training Algorithm}
\paragraph{Pre-training}
Many RL-based models introduce pre-training strategies to refine the agent training efficiency \cite{acl2018qinRL,feng2018RLRE}.
In this study, we pre-train our models in two aspects:
(1) we first pre-train entity and relation extractors to be refined as environment initialization, which is vital to provide reasonable agent states (embeddings of sentences and extracted types).
(2) we then pre-train the policy networks of agents to gain a preliminary ability to evaluate confidence. 
In order to guide the instance confidence evaluation, we extract a small part of the valid data. 
The relatively clean DS type labels of the valid data are used to form states. The binary label is assigned according to the valid data and the policy networks are pre-trained for several epochs.
Although the binary labels from valid data are not exactly the continuous confidence, the policy networks gain a better parameter initialization than random initialization by this approximate training strategy.

\begin{algorithm}[t]
\caption{Training Framework for Extractors}
\label{TrainExtractor}
\begin{algorithmic}[1]
\Require
Noisy training data $D$, pre-trained entity extractor $\theta_e'$, pre-trained relation extractor $\theta_r'$

\Ensure
refined entity/relation extractor $\theta_e^*$, $\theta_r^*$

\State pre-train policy networks of agents based on $\theta_e'$ and $\theta_r'$
\State init: best $F1_e^* \gets F1(\theta_e')$, best $F1_r^* \gets F1(\theta_r')$
\For{epoch $i = 1 \to N$}
\State init: current extractors parameters $\theta_{e} \gets \theta_e'$, $\theta_{r} \gets  \theta_r'$
\For{batch $d_i \in D$}
\State extractors generate $\bf{S^e}$/$\bf{S^r}$ as Equ. (\ref{equ:states})
\State agents take actions (confidences)
\State redistribute instances with confidences
\State train $\theta_{e}$/$\theta_{r}$ with scaled losses $\ell'_e$/$\ell'_r$
\State calculate rewards $r_e$ and $r_r$ as Equ. (\ref{equ:reward})
\EndFor

\IIf{$F1(\theta_e)>F1_e^*$} $F1_e^* \gets F1(\theta_e), \theta_e^* \gets \theta_e$
\IIf{$F1(\theta_r) > F1_r^*$} $F1_r^* \gets F1(\theta_r), \theta_r^* \gets \theta_r$
\EndFor
\end{algorithmic}
\end{algorithm}

\paragraph{Iterative Re-training}
With the pre-trained extractors and policy networks, we retrain extractors and agents as Algorithm \ref{TrainExtractor} detailed.
The agents refine extractors in each epoch and we record parameters of extractors that achieve best F1 performance.
For each data batch, entity and relation extractor perform extraction, form the states $\bf{S^e}$ and $\bf{S^r}$ as Equation (\ref{equ:states}), and send them to entity and relation agents respectively.
Then agents take actions (evaluate confidences) and redistribute instance based on confidences consensus module (Section \ref{sec:consensus}). 
Finally extractors are trained with confidences and give rewards as Equation (\ref{equ:reward}).

\paragraph{Curriculum Learning for Multiagents}
It is difficult to learn from scratch for many RL agents.
In this study, we extend the curriculum learning strategy \cite{bengio2009curriculum} to our cooperative multiagents.
The motivation is that we can leverage the complementarity of the two tasks and enhance the agent exploration by smoothly increasing the policy difficulty.
To be more specific, we maintain a priority queue and sample instances ordered by their reward values.
Once the reward of current sentence excesses the training reward threshold $r_{threshold}$ or the queue is full, we then learn agents policies using Proximal Policy Optimization (PPO) \cite{schulman2017proximal} algorithm, which achieves good performances in many continuous control tasks. 
Algorithm \ref{algo:curriculum} details the training procedure.
\begin{algorithm}[t]
\caption{Curriculum Training with PPO for each Agent}
\label{algo:curriculum}
\begin{algorithmic}[1]
\Require
Data batch $d_i$, queue size $l$, pre-trained policy network with parameter $\Theta'$
\Ensure
Policy network parameter $\Theta$
\State initialize an empty priority queue $q$ with size $l$  
\For{sentence $s_j \in d_i$}
\If{$r_c > r_{threshold}$ or $q$ is full}
\State run policy $\Theta'$ on environment $s_j$
\State compute advantage estimate $\hat{A}$ using Generalized Advantage Estimator (GAE) \cite{schulman2015high}
\State optimize agent loss $\mathcal{L}$ (adaptive KL penalty form) w.r.t $\Theta$ using SGD
\State $\Theta' \gets \Theta$ 
\If{$q$ is full}
\State pull highest priority sentence
\EndIf
\Else
\State insert $s_j$ into $q$ with priority $r_c$
\EndIf
\EndFor

\end{algorithmic}
\end{algorithm}

\section{Experiments}
\subsection{Experimental Setup}
\paragraph{Datasets}
We evaluate our approach on two public datasets used in many extraction studies \cite{pyysalo2007bioinfer,ling2012fine,ren2017cotype}:
\textbf{Wiki-KBP}: the training sentences are sampled from 
Wikipedia articles and the test set are manually annotated from 2013 KBP slot filling task;
\textbf{BioInfer}: the dataset is sampled and manually annotated from biomedical paper abstracts.
The two datasets vary in domains and scales of type set, detailed statistics are shown in Table \ref{tab-data}.

\begin{table}[b]
\centering
\small
\begin{tabular}{|l|l|l|}
\hline
\textbf{Datasets} & \textbf{Wiki-KBP} & \textbf{BioInfer} \\ \hline
\#Relation / entity types & 19 / 126 & 94 / 2,200 \\ \hline
\#Train $M^r$~/~$M^e$ & 148k / 247k & 28k / 53k \\ \hline
\#Test $M^r$~/~$M^e$ & 2,948 / 1,285 & 3,859 / 2,389 \\ \hline
\end{tabular}
\caption{Datasets statistics. $M^r$ and $M^e$ indicates relation and entity mentions respectively.}
\label{tab-data}
\end{table}

\begin{table*}[htp]
\centering
\small
\begin{tabular}{|c|ccc|ccc|}
\hline
\textbf{} & \multicolumn{3}{c|}{\textbf{Wiki-KBP}} & \multicolumn{3}{c|}{\textbf{BioInfer}} \\ \hline
\textbf{Methods} & \textbf{S-F1} & \textbf{Ma-F1} & \textbf{Mi-F1} & \textbf{S-F1} & \textbf{Ma-F1} & \textbf{Mi-F1} \\ \hline
HYENA & 0.26 & 0.43 & 0.39 & 0.52 & 0.54 & 0.56 \\
FIGER & 0.29 & 0.56 & 0.54 & 0.69 & 0.71 & 0.71 \\
WSABIE & 0.35 & 0.55 & 0.50 & 0.64 & 0.66 & 0.65 \\
PLE & 0.37 & 0.57 & 0.53 & 0.70 & 0.71 & 0.72 \\ \hline
CoType & 0.39 & 0.61 & 0.57 & 0.74 & 0.76 & 0.75 \\ 
\multirow{2}{*}{\begin{tabular}[c]{@{}c@{}}\textbf{MRL-CoType} \\ ( improvements)\end{tabular}} & \textbf{0.42}$\pm$7.2e-3 & \textbf{0.64}$\pm$1.1e-2 & \textbf{0.60}$\pm$8.3e-3 & \textbf{0.77}$\pm$6.5e-3& \textbf{0.79}$\pm$1.3e-2& \textbf{0.78}$\pm$7.4e-3\\
 & (+7.69\%) & (+4.92\%) & (+5.26\%) & (+4.05\%) & (+3.95\%) & (+4.00\%) \\ \hline
\end{tabular}
\caption{NER performance on two datasets, 3-time average results with standard deviations are reported.}
\label{tab-entity}
\end{table*}

\begin{table*}[htp]
\centering
\small
\begin{tabular}{|c|ccc|ccc|}
\hline
\textbf{} & \multicolumn{3}{c|}{\textbf{Wiki-KBP}} & \multicolumn{3}{c|}{\textbf{BioInfer}} \\ \hline
\textbf{Methods} & \textbf{Precision} & \textbf{Recall} & \textbf{F1} & \textbf{Precision} & \textbf{Recall} & \textbf{F1} \\ \hline
MintZ & 0.296 & 0.387 & 0.335 & 0.572 & 0.255 & 0.353 \\
MultiR & 0.325 & 0.278 & 0.301 & 0.459 & 0.221 & 0.298 \\
DS-Joint & 0.444 & 0.043 & 0.078 & 0.584 & 0.001 & 0.002 \\
FCM & 0.151 & \textbf{0.500} & 0.301 & 0.535 & 0.168 & 0.255 \\
ARNOR & 0.453  & 0.338 & 0.407 & 0.589 & 0.382 & 0.477 \\
BA-Fix-PCNN & 0.457  & 0.341 & 0.409 & 0.587 & 0.384 & 0.478 \\
RRL-PCNN & 0.435 & 0.322 & 0.392 & 0.577 & 0.381 & 0.470 \\
 \hline
PCNN & 0.423 & 0.310 & 0.371 & 0.573 & 0.369 & 0.461 \\ 
\multirow{2}{*}{\begin{tabular}[c]{@{}c@{}}\textbf{MRL-PCNN} \\ (improvements)\end{tabular}} & \textbf{0.461}$\pm$2.5e-3 & 0.325$\pm$2.3e-3 & 0.407$\pm$1.4e-3 & 0.590$\pm$1.1e-3 & 0.386$\pm$2.3e-3 & 0.483$\pm$2.8e-3\\
 & \multicolumn{1}{c}{(+8.98\%)} & \multicolumn{1}{c}{(+4.83\%)} & \multicolumn{1}{c|}{(+9.70\%)} & \multicolumn{1}{c}{(+2.97\%)} & \multicolumn{1}{c}{(+4.61\%)} & \multicolumn{1}{c|}{(+4.77\%)} \\\hline
CoType & 0.348 & 0.406 & 0.369 & 0.536 & 0.424 & 0.474 \\ 
\multirow{2}{*}{\begin{tabular}[c]{@{}c@{}}\textbf{MRL-CoType} \\ (improvements)\end{tabular}} & 0.417$\pm$1.9e-3 & 0.415$\pm$1.6e-3 & \textbf{0.416}$\pm$1.7e-3 & \textbf{0.595}$\pm$2.1e-3 & \textbf{0.437}$\pm$1.8e-3 & \textbf{0.498}$\pm$2.0e-3\\
 & \multicolumn{1}{c}{(+19.83\%)} & \multicolumn{1}{c}{(+2.22\%)} & \multicolumn{1}{c|}{(+12.74\%)} & \multicolumn{1}{c}{(+11.01\%)} & \multicolumn{1}{c}{(+3.01\%)} & \multicolumn{1}{c|}{(+5.63\%)} \\ \hline
\end{tabular}
\caption{End-to-end relation extraction performance, 3-time average results with standard deviations are reported.}
\label{tab-relation}
\end{table*}

\paragraph{Baselines}
For relation extraction, we compare with both pipe-lined methods and joint extraction methods:
\textbf{MintZ} \cite{mintz2009distant} is a feature-based DS method using a logistic classifier;
\textbf{MultiR} \cite{hoffmann2011knowledge} models noisy DS labels with multi-instance multi-label learning;
\textbf{DS-Joint} \cite{li2014incremental} jointly extracts entities and relations using structured perceptron;
\textbf{FCM} \cite{gormley2015improved} introduces a neural model to learn linguistic compositional representations;
\textbf{PCNN} \cite{zeng2015distant} is an effective relation extraction architecture with piece-wise convolution;
\textbf{CoType} \cite{ren2017cotype} is a state-of-the-art joint extraction method leveraging representation learning for both entity and relation types;
\textbf{RRL-PCNN} \cite{acl2018qinRL} is a state-of-the-art RL-based method, which takes PCNN as base extractor and can also be a plugin to apply to different relation extractors;
\textbf{ARNOR} \cite{jia2019arnor} is a state-of-the-art de-noising method, which proposes attention regulation to learn relation patterns;
\textbf{BA-fix-PCNN} \cite{ye2019looking} greatly improves the extraction performance by introducing 20\% samples of the test set and estimate its label distribution to adjust the classifier of PCNN.

For entity extraction methods, we compare with a supervised type classification method, \textbf{HYENA} \cite{yosef2012hyena}; a heterogeneous partial-label embedding method, \textbf{PLE} \cite{ren2016label}; and two DS methods \textbf{FIGER} \cite{ling2012fine} and \textbf{WSABIE} \cite{yogatama2015embedding}.

\paragraph{Multiagents Setup}
To evaluate the ability of our approach to refine existing extractors, we choose two basic extractors for our \textbf{M}ultiagent \textbf{RL} approach, CoType and PCNN, and denote them as \textbf{MRL-CoType} and \textbf{MRL-PCNN} respectively. 
Since PCNN is a pipe-lined method, we reuse a pre-trained and fixed CoType entity extractor, and adopt PCNN as base relation extractor to adapt to the joint manner.
For the CoType, we use the implementation of the original paper \footnote{https://github.com/INK-USC/DS-RelationExtraction}, and
adopt the same sentence dimension, type dimension and hyper-parameters settings as reported in \cite{ren2017cotype}.
For the PCNN, we set the number of kernel to be 230 and the window size to be 3.
For the KG embeddings, we set the dimension to be 50 and pre-train them by TransE.
We use Stochasitc Gradient Descent and learning rate scheduler with cosine annealing to optimize both the agents and extractors, the learning rate range and batch size is set to be [1e-4, 1e-2] and 64 respectively.

We implement our RL agents using a scalable RL library, RLlib \cite{liang2018rllib}, and adopt 2/8 relation agents and 2/16 entity agents for Wiki-KBP/BioInfer datasets respectively, according to their scales of type sets. 
For the multi-agents, due to the limitation of RL training time, we set the PPO parameters as default RLlib setting and perform preliminary grid searches for other parameters.
For the PPO algorithm, we set the GAE lambda parameter to be 1.0, the initial coefficient for KL divergence to be 0.2. 
The loss adjusting factor $\lambda$ is searched among \{1, 2, 4\} and set to be 2, the reward control factors $\alpha$ is searched among \{2e-1, 1, 2, 4\} and set to be 2.
For all agents, the dimensions of GRU is searched among \{32, 64\}, 
and the setting as 64 achieved sightly better performance than setting as 32, while the larger dimension setting leads to higher memory overhead for each agent.
Hence we set it to be 32 to enable a larger scale of the agents.

\begin{table*}[htp]
\centering
\small
\begin{tabular}{|c|ccc|ccc|} 
\hline
\textbf{} & \multicolumn{3}{c|}{\textbf{Wiki-KBP}} & \multicolumn{3}{c|}{\textbf{BioInfer}} \\ \hline
\textbf{Settings} & \textbf{Precision(\%)} & \textbf{Recall(\%)} & \textbf{F1(\%)} & \textbf{Precision(\%)} & \textbf{Recall(\%)} & \textbf{F1(\%)} \\ \hline
Curriculum & 41.7$\pm$0.19    & 41.5$\pm$0.16  & 41.6$\pm$0.17 & 59.5$\pm$0.21    & 43.7$\pm$0.18  & 49.8$\pm$0.20   \\
Joint (w/o curriculum)          &                                41.3$\pm$0.22    & 40.9$\pm$0.20  & 41.1$\pm$0.21  & 58.7$\pm$0.24    & 42.6$\pm$0.19  & 48.5$\pm$0.23  \\ 
Separate  (w/o joint)    &                         38.8$\pm$0.24    & 40.5$\pm$0.27  & 38.4$\pm$0.25   & 54.7$\pm$0.27    & 41.3$\pm$0.23  & 47.6$\pm$0.26 
\\  \hline
\end{tabular}
\caption{Ablation results of the MRL-CoType for end-to-end relation extraction.}
\label{tab-ablation}
\end{table*}

\subsection{Effectiveness of Multiagents}
\subsubsection{Performance on Entity Extraction}
We adopt the Macro-F1, Micro-F1 and Strict-F1 metrics \cite{ling2012fine} in the entity extraction evaluation. For Strict-F1, the entity prediction is considered to be ``strictly'' correct if and only if when the true set of entity tags is equal to the prediction set. 
The results are shown in Table \ref{tab-entity} and we can see that our approach can effectively refine the base extractors and outperform all baseline methods on all metrics.
Note that the refinements on BioInfer is significant (t-test with $p<0.05$) even though the BioInfer has a large entity type set (2,200 types) and the base extractor CoType has achieved a high performance (0.74 S-F1), which shows that our agents are capable of leading entity extractors towards a better optimization with noisy.

\subsubsection{Performance on Relation Extraction}
Another comparison is the end-to-end relation extraction task, we report the precision, recall and F1 results in Table \ref{tab-relation} and it illustrates that: 

(1) Our method achieves best F1 for Wiki-KBP, outperforms all baselines on all metrics for BioInfer data, and significantly refines both the two base extractors, PCNN and CoType (t-test with $p<0.05$), demonstrating the effectiveness of our approach.

(2) The improvements for CoType are larger than PCNN. Since CoType is a joint extraction model and leverages multi-agents better than the single-task extractor with fixed entity extractor. This shows the benefit of correlations between the two extraction tasks.

(3) Using the same base relation extractor, the MRL-PCNN achieves significantly better improvements than RRL-PCNN (t-test with $p<0.05$). 
Besides, the precision of RRL-PCNN method is relatively worse than recall, which is mainly caused by the noise propagation of entity extraction and its binary discard-or-retain action.
By contrast, our model achieves better and more balanced results by leveraging the cooperative multiagents with fine-grained confidences.

(4) The MRL-PCNN gains comparable performance with BA-Fix-PCNN, which leverages the additional information from the test set to adjust softmax classifier. This verifies the effectiveness and the robustness of the proposed RL-based re-labeling method to reduce the shifted label distribution gap without knowing the test set.

\subsection{Ablation Analysis}
To evaluate the impact of curriculum learning strategy and joint learning strategy of our method, we compare three training settings: 
\textbf{curriculum learning}, standard training procedure as described in Section \ref{sec:training};
\textbf{joint multiagents training} without curriculum learning (randomly sample training instances);
and \textbf{separate training} without the participation of other agents using a pipeline manner, i.e., train an entity agent with only entity extractor and train a relation agent with only relation extractor.

\begin{figure}[tb]
\centerline{\includegraphics[width=0.85
\linewidth]{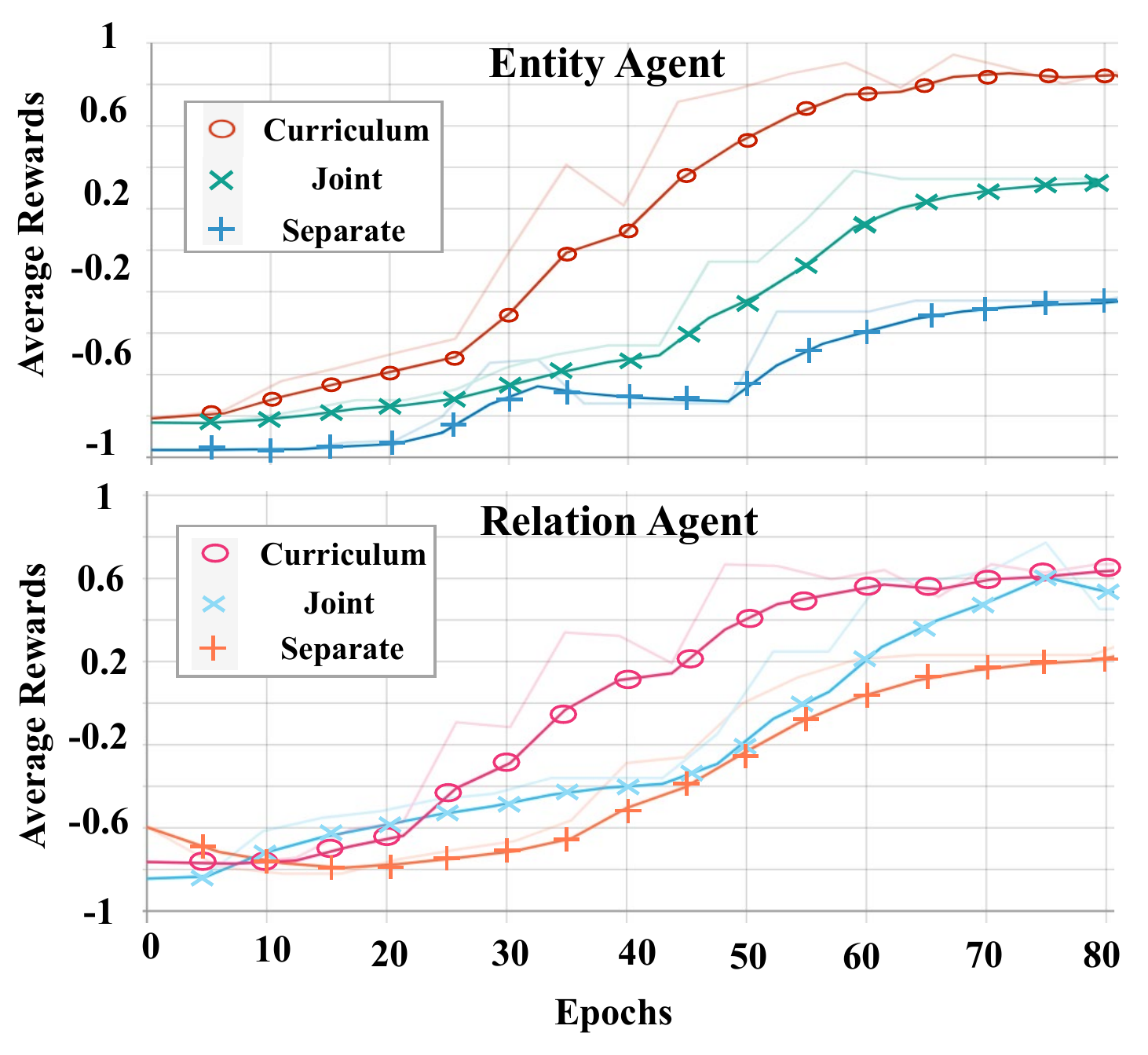}}
\caption{Smoothed average rewards on Wiki-KBP data for two agents of MRL-CoType. The light-colored lines are un-smoothed rewards. }
\label{fig.relation-KBP}
\end{figure}

The end-to-end relation extraction results are reported in Table \ref{tab-ablation}.
The curriculum setting and the joint setting achieve much better results than the separate training setting. This shows the superiority of cooperative multi-agents over single view extraction, which evaluates confidences with limited information.
Besides, the curriculum setting achieves better results than the joint setting, especially on the BioInfer data, which has a larger type set and is more challenging than Wiki-KBP. This indicates the effectiveness of the curriculum learning strategy, which enhances the model ability to handle large state space with gradual exploration.

Training efficiency is an important issue for RL methods since the agents face the exploration-exploitation dilemma. 
We also compare the three settings from the view of model training. 
Figure \ref{fig.relation-KBP} reports the average rewards for an entity agent and a relation agent on Wiki-KBP respectively. 
A high average reward indicates that the agent is trained effectively since it made valuable decisions and received positive feedback.
From it we have the following observations: 
(1) The curriculum setting and the joint setting gain better performance than the separate training, which is consistent with the end-to-end extraction results. 
The improvement comes from the mutual enhancement among agents, since the correlations between the two tasks can restrict the reward signals to only those agents involved in the success or failure on the task;
(2) The curriculum learning achieves higher rewards than the other two settings with fewer epochs,
since that the convergence to local optimum can be accelerated by smoothly increasing the instance difficulty, and the multiagents provide a regularization effect.

\begin{figure}[tb]
\centerline{\includegraphics[width=0.99
\linewidth]{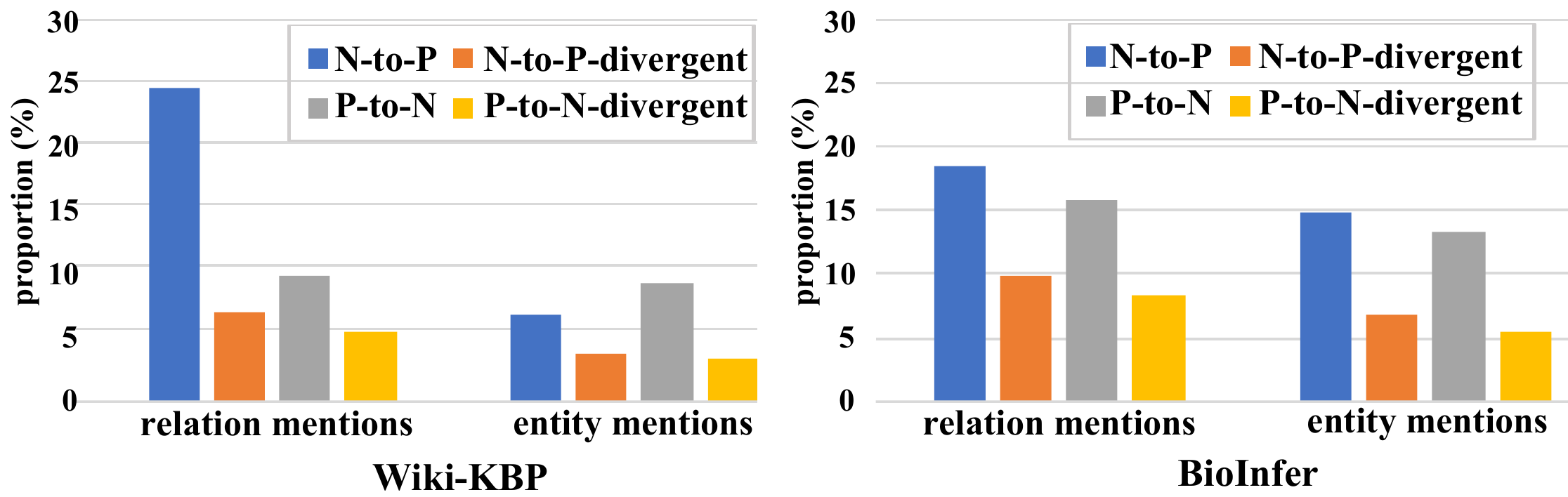}}
\caption{Proportions of re-labeled instances for MRL-CoType. ``N-to-P'' denotes the instances are re-labeled from negative to positive. ``divergent'' means that entity agents and relation agent have different evaluations about whether the instance is positive or negative.}
\label{fig.relabel-stat}
\end{figure}

\begin{table*}[tb]
\centering
\resizebox{\textwidth}{!}{%
\begin{tabular}{|l|l|l|l|l|l|}
\hline
\hline
 \begin{tabular}[c]{@{}l@{}}Sentence 1, False Negative, \\ Label: (\textit{Bashardost}{[}/person{]}, \textbf{None}, \textit{Ghazni}{[}/location{]})\end{tabular} & Entity Extractor & \begin{tabular}[c]{@{}l@{}}Relation \\ Extractor\end{tabular} & \begin{tabular}[c]{@{}l@{}}Entity \\ Agents\end{tabular} & \begin{tabular}[c]{@{}l@{}}Relation \\ Agent\end{tabular} & \begin{tabular}[c]{@{}l@{}}Confidence \\ Consensus\end{tabular} \\ \hline
\begin{tabular}[c]{@{}l@{}}\textit{Bashardost}, an ethnic Hazara, was born in \textit{Ghazni} \\ province to a family of government employees.\end{tabular} & \begin{tabular}[c]{@{}l@{}}\textit{Bashardost}{[}/person{]}\\ \textit{Ghazni}{[}/location{]}\end{tabular} & \begin{tabular}[c]{@{}l@{}}\textbf{country}\_\\ \textbf{of\_birth}\end{tabular} & \begin{tabular}[c]{@{}l@{}}0.772\\ 0.729\end{tabular} & \begin{tabular}[c]{@{}l@{}}0.896\\ \end{tabular} & \begin{tabular}[c]{@{}l@{}}Positive\\ (0.799)\end{tabular} \\
 \hline \hline
\begin{tabular}[c]{@{}l@{}}Sentence 2, False Positive, Label: (\textit{profilin}{[}/\textbf{Protein}{]}, \\ \textbf{ POS\_ACTION\_Physical}, \textit{actin}{[}/Protein{]})\end{tabular}
 &  &  &  &  &  \\ \hline
 \begin{tabular}[c]{@{}l@{}} Acanthamoeba \textit{profilin} affects the mechanical \\ properties of nonfilamentous \textit{actin}.\end{tabular} &
 \begin{tabular}[c]{@{}l@{}}\textit{profilin}{[}/\textbf{None}{]} \\ \textit{actin}{[}/Protein{]}\end{tabular} & \textbf{None} & \begin{tabular}[c]{@{}l@{}}0.373\\ 0.791\end{tabular} & \begin{tabular}[c]{@{}l@{}}0.236\\ \end{tabular} & \begin{tabular}[c]{@{}l@{}}Negative\\ (0.533)\end{tabular} \\
 \hline
\end{tabular}%
}
\caption{Confidence evaluations on two noisy instances using MRL-CoType.}
\label{tab:case-study}
\end{table*}

\subsection{Re-labeling Study}
To gain insight into the proposed method, we conduct a statistic on the final re-labeled instances.
Figure \ref{fig.relabel-stat} reports the results and shows that our approach identifies some noisy instances including both positives and negatives, and leverage them in a fine-grained manner comparing with discard-or-retain strategy.
Besides, the instances which are re-labeled from negatives to positives take a larger proportion than those with inverse re-labeling assignments, especially on Wiki-KBP data. 
This is in accordance with the fact that many noisy labels are ``None'' in DS setting.
Note that some instances are re-labeled with divergent evaluations between entity-view and relation-view agents, which are usually get low confidences through the consensus module and have a small impact on the optimization with damping losses. 

We further sample two sentences to illustrate the re-labeling processes.
On Table \ref{tab:case-study}, the first sentence has a noisy relation label \textbf{None}, while the relation extractor recognizes it as \textbf{country\_of\_birth} relation.
Based on the extracted type, the relation-view agent evaluates it as a confidential positive instance due to the typical pattern ``born in'' in the sentence.
The entity-view agents also evaluate it as positive with relatively lower confidences,
and finally the sentence is re-labeled as positive by the consensus module. 
For the second sentence, agents disagree that it is positive.
With the help of diverse extraction information, the consensus module re-labels the instance with low confidence score, and further alleviates the performance harm by loss damping.

\section{Related Works}
Many entity and relation extraction methods have been proposed with the pipelined fashion, i.e., perform named entity recognition (NER) first and then relation classification.
Traditional NER systems usually focus on a few predefined types with supervised learning \cite{yosef2012hyena}.
However, the expensive human annotation blocks the large-scale training data construction.
Recently, several efforts on DS and weak supervision (WS) NER extraction have been made to address the training data bottleneck \cite{yogatama2015embedding,yang2018-distantly}.
For relation extraction, there are also many DS methods \cite{mintz2009distant,min2013distant,zeng2015distant,han2016global,ji2017distant,lei2018cooperative} and WS methods \cite{jiang2009multi,ren2016label, deng2019medtruth} to address the limitation of supervised methods.
Our method can be applied for a large number of those extractors as a post-processing plugin since the DS and WS usually incorporate many noises.

A recent work CrossWeigh \cite{wang2019crossweigh} estimates the label mistakes and adjusts the weights of sentences in the NER benchmark CoNLL03. They focus on the noises of supervised ``gold standard'' labels while we focus on the noises of automatically constructed ``silver standard'' labels.
Moreover, we deal with the noises by considering the shifted label distribution problem, which is overlooked by most existing DS works. In \citet{ye2019looking}, this issue is analyzed and authors improve performance significantly by using the distribution information from test set. 
In this paper, we propose to use RL to explore suitable label distributions by re-distributing the training set with confidence-scored labels, which is practical and robust to label distribution shift since we may not know the distribution of test set in real-world applications.

Another extraction manner is joint extraction, such as methods based on neural network with parameter sharing \cite{miwa2016end}, representation learning \cite{ren2017cotype} and new tagging scheme \cite{zheng2017joint}. However, these works perform extraction without explicitly handling the noises.
Our approach introduces multiagents to the joint extraction task and explicitly model sentence confidences.
As for the RL-based methods, in \citet{zeng2018large}, RL agent is introduced as bag-level relation predictor. \citet{acl2018qinRL} and \citet{feng2018RLRE} use agent as instance selectors to discard noisy instances in sentence-level.
Different from adopting a binary action strategy and only focus on false positives in these works, we adopt a continuous action space (confidence evaluation) and handle the noises in a fine-grained manner.
The binary selection strategy is also adopted in a related study, Reinforced Co-Training \cite{wu2018reinCo}, which uses an agent to select instances and help classifiers to form auto-labeled datasets. 
An important difference is that they select unlabeled instances while we evaluate noisy instances and re-label them.
More recently, HRL \cite{takanobu2019hierarchical} uses a hierarchical agent to first identifies relation indicators and then entities. 
Different from using one task-switching agent of this work, we leverage a group of multiagents, which can be a pluggable helper to existing extraction models.

\section{Conclusions}
To deal with the noise labels and accompanying shifted label distribution problem in distant supervision, in this paper, we propose a novel method to jointly extract entity and relation through a group of cooperative multiagents.
To make full use of each instance, each agent evaluates the instance confidence from different views, and then a confidence consensus module is designed to re-label noisy instances with confidences.
Thanks to the exploration of suitable label distribution by RL agents, the confidences are further used to adjust the training losses of extractors and the potential harm caused by noisy instances can be alleviated.
To demonstrate the effectiveness of the proposed method, we evaluate it on two real-world datasets and the results confirm that the proposed method can significantly improve extractor performance and achieve effective learning.

\section*{Acknowledgements}
This work is supported by the National Natural Science Foundation of China (No.61602013), and the Shenzhen General Research Project (No. JCYJ20190808182805919).

\balance

\bibliography{mybib}
\bibliographystyle{acl_natbib}

\end{document}